\documentclass[runningheads]{llncs}

\usepackage{booktabs}
\usepackage{url}
\usepackage{subfigure}
\usepackage{color}
\usepackage{array}
\usepackage{multirow}
\usepackage{multicol}
\usepackage{amsmath}
\usepackage{amssymb}
\usepackage{verbatim}
\usepackage{graphicx}
\usepackage{CJKutf8}
\usepackage{setspace}
\usepackage{bm}
\begin{document}
\begin{CJK}{UTF8}{gbsn}

\title{Distant Supervision for E-commerce Query Segmentation via Attention Network}

%

\author{Zhao Li\inst{1} 
\and Donghui Ding\inst{1} 
\and Pengcheng Zou\inst{1} 
\and Yu Gong\inst{1}
\and Xi Chen\inst{1}
\and Ji Zhang\inst{2}
\and Jianliang Gao\inst{3}
\and Youxi Wu\inst{4}
\and Yucong Duan\inst{5}
}
%

%
\institute{Alibaba Group, China \and
The University of Southern Queensland, Australia \and
Central South University, China \and
Hebei University of Technology,  China \and Hainan University, China
\\
\email{\{lizhao.lz,donghui.ddh,xuanwei.zpc,gongyu.gy\}@alibaba-inc.com; gongda.cx@taobao.com; Ji.Zhang@usq.edu.au, gaojianliang@csu.edu.cn; wuc567@163.com; duanyucong@hotmail.com}}

\maketitle              

\begin{abstract}
The booming online e-commerce platforms demand highly accurate approaches to segment queries that carry the product requirements of consumers. Recent works  have shown that the supervised methods, especially those based on deep learning, are attractive for achieving better performance on the problem of query segmentation. However, the lack of labeled data is still a big challenge for training a deep segmentation network, and the problem of Out-of-Vocabulary (OOV)  also  adversely impacts  the  performance  of  query segmentation. Different from query segmentation task in an open domain, e-commerce scenario can provide external documents that are closely related to these queries. Thus, to deal with the two challenges,  we employ the idea of distant supervision and design a novel method to find contexts in external documents and extract features from these contexts. In this work, we propose a BiLSTM-CRF based model with an attention module to encode external features, such that external contexts information, which can be utilized naturally and effectively to help query segmentation. Experiments on two datasets show the effectiveness of our approach compared with several kinds of baselines.
\end{abstract}
\section{Introduction}
\label{sec:intro}

Query segmentation is an important task in information retrieval (IR).
A query is a sequence of words (in English) or characters (in Chinese) which carries the information requirement of a user.
Query segmentation task is to cut a query into several continuous subsequences called \emph{segments} that are normally frequently-used phases.
Compared to the independent words or characters in the query, these meaningful segments are more significant to the search engine.
Assuming a user is looking for ``short sleeve long dress'', where ``short sleeve'' and ``long dress'' are two segments that indicate a long dress with a short sleeve.
If the query is processed based on independent words, many irrelevant short dresses or clothes with long sleeve may be returned. 
The quality of query segmentation is very important for the downstream IR task.


Query segmentation task has been studied extensively in research community.
The existing methods can be mainly divided into three categories: unsupervised \cite{risvik_query_2003,mishra_unsupervised_2011,parikh_segmentation_2013,hagen_power_2010,hagen_query_2011,tan_unsupervised_2008},
feature-based supervised \cite{yu_query_2009,du_perceptron-based_2014}
and deep learning methods \cite{kale_towards_2017,lin_query_2017}.
Unsupervised methods score each segmentation combination of a query by some kinds of statistical indexes like mutual information \cite{risvik_query_2003}.
Feature-based and deep learning methods are supervised, and rely on a large number of gold segmented queries.

Supervised methods, especially deep learning, are attractive for achieving a better performance and are focused in our work. However, a lack of labeled data is one of big challenges for training deep neural networks.
In this work, we employ the idea of long-distance supervision \cite{mintz2009distant} to automatically create large amounts of gold standard data.
In the e-commerce field, because queries are related to products, we build a dictionary by crawling brand names, product names, attribute names and attribute values from the product detail pages of online shopping platforms. Then, a simple max-matching algorithm can be used to segment queries by matching subsequences in queries with the words in the dictionary.

Another challenge of the query segmentation task is the so-called Out-of-Vocabulary (OOV) segments \cite{varjokallio2016unsupervised,huang2017addressing}.
OOV segments are those segments that are used in the test queries but do not appear in the training queries.
OOV issue has been studied widely in Chinese Word Segmentation (CWS) tasks
\cite{Xue2003Chinese,Jin2005A,Zhao2006An,huang2007chinese} that are very similar to query segmentation.
\cite{huang2007chinese} argues that OOV is the key problem of CWS task.
We indicate that OOV also impacts the performance of query segmentation to a large extent.
We try to alleviate OOV issue by incorporating contexts from external documents.
If the OOV segments can be found in external documents, we can extract some valuable features to help recognize them.

\begin{figure}[th]
	\centering
	\includegraphics[width=0.35\columnwidth]{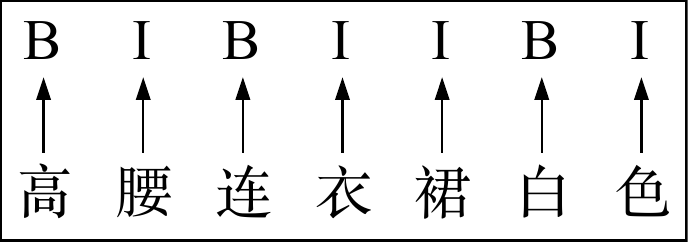}
	\caption{The segmentation result is ``高腰 (high-waisted) / 连衣裙 (dress) / 白色 (white)''. Therefore, the label sequence of this query is ``B/I/B/I/I/B/I''.}
	\label{fig:sq}
\end{figure}

We treat query segmentation as a sequence labeling problem.
The tagging scheme consists of ``B'' and ``I''.
``B'' (Begin) means the current character is the head of its segment,
while ``I'' (Inner) means the current character belongs to the previous segment.
An example is shown in Figure \ref{fig:sq}.
Under this tagging scheme,
the target is to predict one label of ``B'' and ``I'' for each character in a query.
Our approach contains two steps.
The first step is to find contexts for each character in queries and extract features from these contexts.
The second step is to train our neural networks model by using these extracted features.
As for each character in queries, we can Obtain its left and right bi-grams. For example, the left and right bi-grams of character ``衣'' in the query ``高腰连衣裙白色'' are ``连衣'' and ``衣裙'', respectively.

\begin{figure}[th]
	\centering
	\includegraphics[width=1.0\columnwidth]{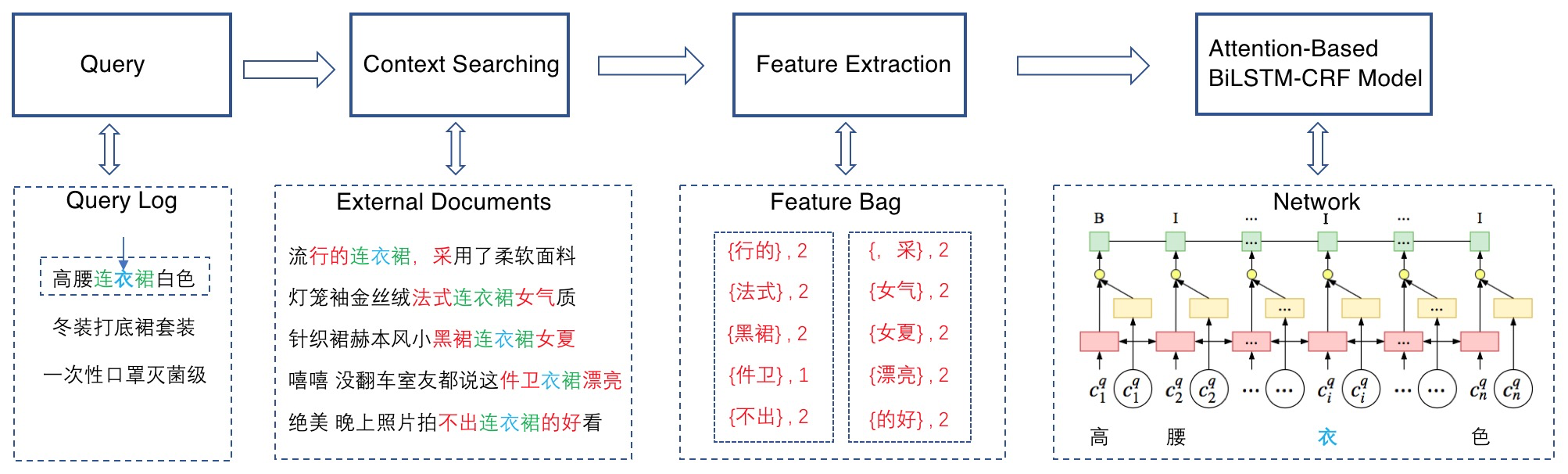}
	\caption{The algorithm contains 3 main parts, i.e., context searching, feature extraction and attention-based BiLSTM-CRF model. In the context searching and feature extraction parts, we utilize external documents to extract features. Then, the query and extracted features are encoded into attention-based BiLSTM-CRF model. The figure demonstrates how to extract the character and distance features of the character``衣''in the query ``高腰连衣裙白色'', where the left bi-gram is ``连衣''and right bi-gram is ``衣裙''.}
	\label{fig:arch}
\end{figure}

An shown in Figure \ref{fig:arch}, we search these two bi-grams in external documents.
All sentences that contain any one of them are treated as the contexts of this character.
All the contexts of a character is called context bag. For each context in the context bag, we use the same method to extract features and get the feature bag.
BiLSTM-CRF \cite{Huang2015Bidirectional,Ma2016End} is a common model to deal with sequence labeling problem.
It can be used for our query segmentation directly.
To utilize the feature bag from external documents,
we design our model by improving the normal BiLSTM-CRF model with attention mechanism \cite{bahdanau2014neural}, which is used to encode the feature bag and produce its vector representation.
Then, we add this vector representation of feature bag to the commonly used BiLSTM-CRF.
The helpful information contained in contexts will help predict character labels.


Our main contributions of this paper are as follows:
\begin{itemize}
	\item We employ the idea of distant supervision method and propose an effective method to label e-commerce queries automatically, which addresses the problem of the lack of labeled data;
	\item We propose a BiLSTM-CRF based model With attention mechanism to contexts information from external documents, which can alleviate the issue of Out-of-Vocabulary (OOV);
	\item Experiments on two e-commerce datasets show that our model can achieve 0.049 and 0.023 improvements in F1 value above the strongest baselines.
\end{itemize}

\section{Contexts and Features}
\label{sec:framework}

Assuming that $\mathcal{Q}$ is the query log, a query $q \in \mathcal{Q}$ is a sequence of characters $q = (c_1^q, c_2^q, \ldots, c_i^q, \ldots, c_n^q)$. The external documents are a plain document set $\mathcal{D}$. For each character $c_i^q$ in $q$, we search in documents $\mathcal{D}$ and find its contexts. All the contexts of $c_i^q$ form a context bag $B_i$ that is a set of sentences. Given the context bag $B_i$, we design a simple but novel method to extract same kinds of features from each context and obtain the feature bag $F_i$.

\subsection{Context Searching}

\begin{figure*}[th]
	\centering
	\includegraphics[width=0.6\columnwidth]{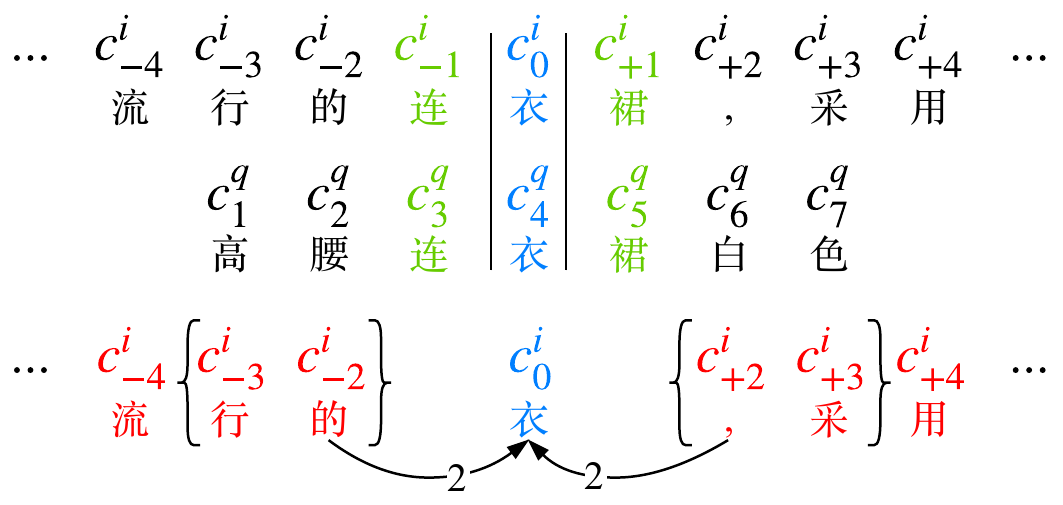}
	\vspace{-1em}
	\caption{This is an example to show the process of extracting features. First line is one of the contexts of character ``衣''. Second line is the query. Third line is the subtraction result.}
	\label{fig:subtract}
\end{figure*}
For character $c_i^q$ in $q$, there are $4$ possible cases about its boundary information: $(1)$ $c_i^q$ is the begin and the end of current segment, which means $c_i^q$ forms an independent segment. The length of this segment is $1$; $(2)$ $c_i^q$ is the begin of current segment but not the end. The length of current segment is more than $1$. What's more, we can infer that the right bi-gram $c_i^qc_{i+1}^q$ of $c_i^q$ should belong to current segment; $(3)$ $c_i^q$ is the end of current segment but not the begin. Also the length of this segment is longer than $1$. Thus, we know that the left bi-gram $c_{i-1}^q c_i^q$ should be in this segment; $(4)$ $c_i^q$ is in the middle position of current segment. This means tri-gram $c_{i_1}^q c_i^q c_{c+1}^q$ should be in this segment. In other words, the left bi-gram is in one segment, and the same for the right bi-gram. For case $(1)$ and $(2)$, the label of $c_i^q$ is ``B'', and for case $(3)$ and $(4)$, the label is ``I''.

From the view of left and right bi-grams, the existence of left bi-gram $c_{i_1}^q c_i^q$ indicates whether $c_i^q$ is the begin of current segment, while the existence of right bi-gram indicates whether $c_i^q$ is the end of current segment. Because we do not know which bi-gram really exists, we search both two bi-grams in external documents $\mathcal{D}$. Sentences that contain any one of these two bi-grams are put into the context bag $B_i$. If a bi-gram exists, this bi-gram should be used frequently in documents. Therefore, we can find many contexts that can support this bi-gram. Conversely, if a bi-gram does not exist, there should be few contexts which support this bi-gram. For example, in the query of ``高腰 / 连衣裙 / 白色'', the two bi-gram of the character ``腰'' are ``高腰'' and ``腰连''. Because ``高腰'' is a common word in the dress category, there are many contexts containing segment ``高腰'' in documents $\mathcal{D}$. ``腰连'' is not a common word in Chinese, so few contexts can be found to support ``腰连''. Further, we can make a conclusion that ``腰'' is case $(3)$. Note that there is no need to judge whether a context is found by the left or right bi-gram. The distribution of contexts in context bag will decide the existence of these two bi-grams.


\subsection{Feature Extraction}

As we have mentioned, the two nearest bi-grams of $c_i^q$ are related to its boundary information and are used to search the contexts. We can also use the existence of these bi-grams in each contexts as boundary information. For example, a pair of Boolean number $(1, 0)$ can be extract from a context. Number $1$ in the pair means the left bi-gram is mentioned in this context, while the $0$ in the pair indicates the right bi-gram is not mentioned. However, we design a novel method to extract the boundaries of $c_i^q$ in $q$, which is much more informative than the Boolean pair.

The context bag $B_i$ is a set of sentences $(s_1, s_2, \ldots, s_j, \ldots)$. Each context must contain at least one of two bi-grams, $c_{i-1}^q c_i^q$ or $c_i^q c_{i+1}^q$. Therefore, $c_i^q$ must appear in any of these contexts. For each context, we treat $c_i^q$ as the center character, and apply same method to extract features. We use context $s_j$ in $B_i$ as an example to illustrate how to extract the features. $s_j$ is a sequence of characters $(\ldots, c_{-3}^j, c_{-2}^j, c_{-1}^j, c_{0}^j, c_{+1}^j, c_{+2}^j, c_{+3}^j, \ldots)$ where $c_0^j$ is the center character, which is $c_i^q$. Characters in the left of $c_0^j$ is called the left part, while characters in the right of $c_0^j$ is the right part. We first align $s_j$ with the query $q$ according to $c_0^j$ in $s_j$ and $c_i^q$ in $q$. Then, we subtract $q$ from $s_j$ character by character, which means we go through $s_j$ from center to two sides and take away the shared characters with $q$. Assuming the difference between $s_j$ and $q$ is $(\ldots, c_{-k_l-1}^j, c_{-k_l}^j, c_{0}^j, c_{+k_r}^j, c_{+k_r+1}^j, \ldots)$, where both characters between $c_{-k_l}^j$ and $c_{0}^j$ in left part and characters between $c_{+k_r}$ and $c_{0}^j$ in right part are taken away. Figure \ref{fig:subtract} shows this process. The first line is a context of character ``衣''. The second line is the query. We align the context with query by ``衣''. Characters ``连'' and ``裙'' are taken away. And both $k_l$ and $k_r$ are $2$.

The left part of difference is used to extract features about the left boundary of current segment, while the right part is for the right boundary. Using the left part as an example, $k_l$ can be treated as the distance between the left boundary and the center character. In Figure \ref{fig:subtract}, $k_l$ is 2, which means there are $2$ characters (including the center character itself) between the left boundary and the center character ``衣''. Additionally, characters near the left boundary can also help to support the current segment. When the window size is $2$, the left character features are $\{c_{-k_l-1}^j, c_{-k_l}^j \}$. In Figure \ref{fig:subtract}, left character features are $\{$``行", ``的" $\}$ where ``的'' is a typical stop signal in Chinese. As a conclusion, for the left boundary, there are two kinds of features, i.e., the distance and character features. As for context $s_j$ of $c_i^q$, left feature $L_j$ is $(\{ c_{-k_l-1}^j, c_{-k_l}^j \}, k_l)$. Similarly, we can have the right feature $R_j=(\{ c_{+k_r}^j, c_{+k_r+1}^j \}, k_r)$ that is for the right boundary.

By applying above processes to every context in context bag $B_i$, we can get the feature bag $F_i = (\left \langle L_1, R_1 \right \rangle, \left \langle L_2, R_2 \right \rangle, \cdots, \left \langle L_j, R_j \right \rangle, \cdots)$ where $\left \langle L_j, R_j \right \rangle$ is the features from context $s_j$. $F_i$ will be used to help to predict the label of $c_i^q$.






\section{Networks}
\label{sec:methodology}

\begin{figure*}[h!t]
\centering
\subfigure[Legend]{\label{fig:11}\includegraphics[scale=0.5]{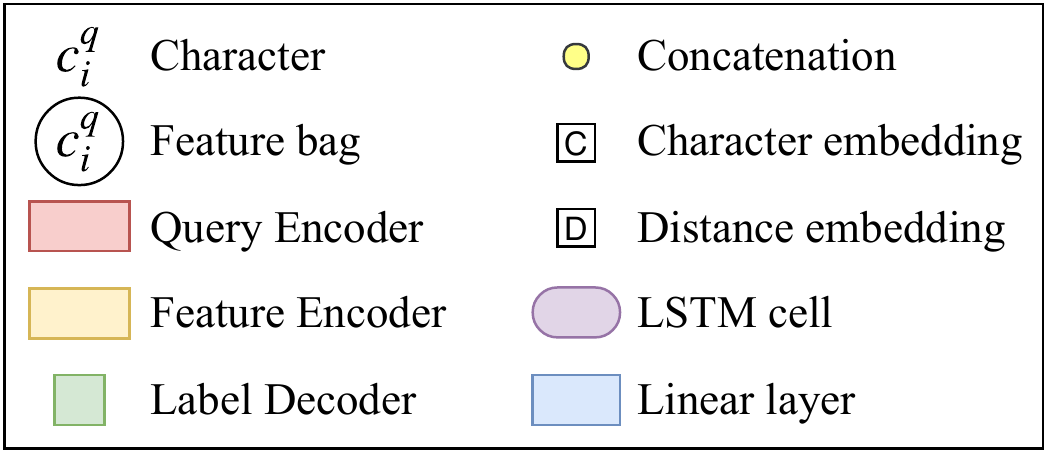}}
\subfigure[Architecture]{\label{fig:112}\includegraphics[scale=0.5]{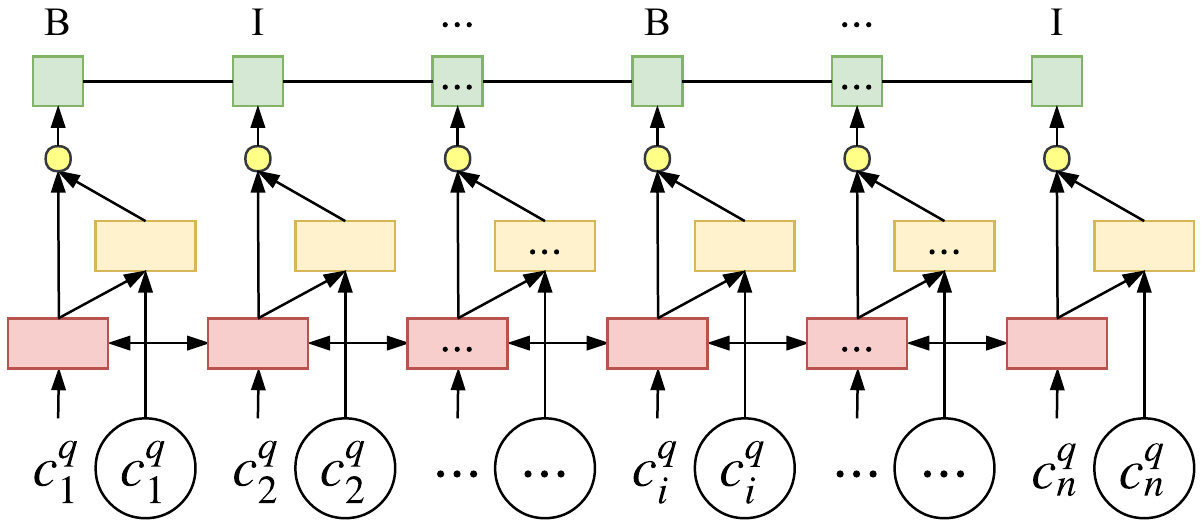}}
\subfigure[Query Encoder]{\label{fig:21}\includegraphics[scale=0.5]{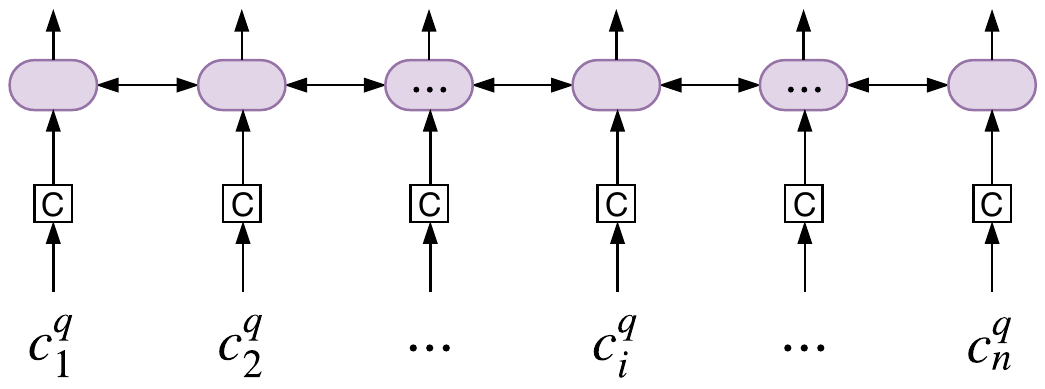}}
\subfigure[Feature Encoder]{\label{fig:22}\includegraphics[scale=0.5]{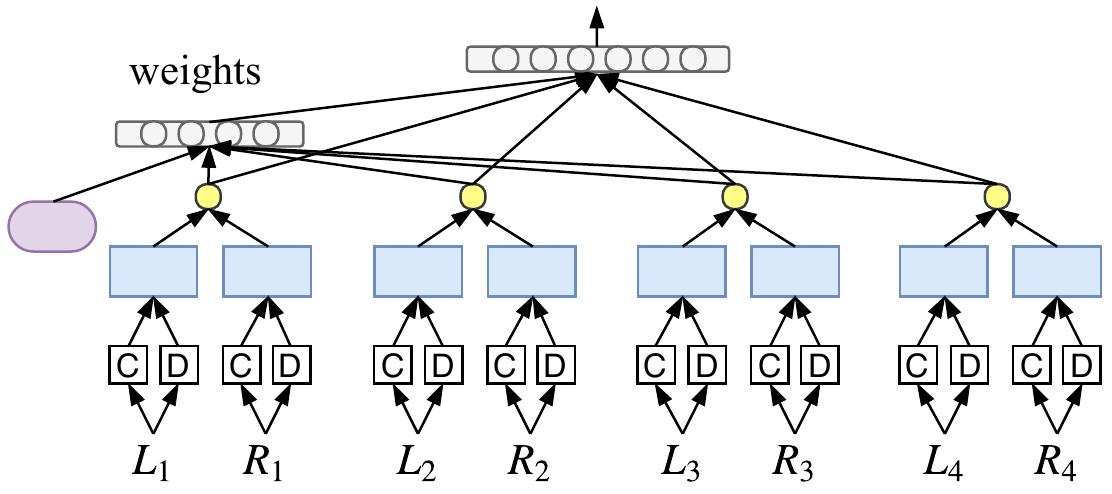}}
\vspace{-1em}
\caption{There are 4 sub-figures, Legend, Architecture, Query Encoder and Feature Encoder. (a) shows all legends used in other 3 sub-figures. (b) is the architecture of our model which contains three module, Query Encoder, Feature Encoder and Label Decoder. (c) shows the detail of query encoder module which is a LSTM structure. (d) shows the detail of feature encoder module which is based on the attention mechanism.}
\label{fig:model}
\end{figure*}

Given a query $q = (c_1^q, \ldots, c_i^q, \ldots, c_n^q)$, our model predicts the label sequence $\bm{y} = (y_1, \ldots, y_i, \ldots, y_n)$, where $y_i$ is the label of character $c_i^q$. BiLSTM-CRF is a typical deep learning model to deal with sequence labeling tasks. BiLSTM-CRF can be used for query segmentation task directly. The input to BiLSTM-CRF is a sequence of characters in $q$. Then, a BiLSTM module encodes this character sequence, and get a feature sequence. The final CRF module will decode the label sequence $\bm{y}$ based on the feature sequence. In the BiLSTM-CRF model, label $y_i$ of $c_i^q$ only relies on the characters in $q$. Now, for each character $c_i^q$ we extract a feature bag $F_i$ from external documents that can provide helpful boundary information. To take advantage of $F_i$, we extend the BiLSTM-CRF model by adding an additional module which takes $F_i$ as another input. The feature bag $F_i$ is extracted from several contexts. We argue that the boundary information carried by different contexts is not equally important. Therefore, this added module uses attention mechanism to deal with this difference. Figure \ref{fig:model} shows the architecture of our model, which consists of $3$ modules, i.e., a query encoder, a feature encoder and a label decoder.

\subsection{Query Encoder}

As shown in Figure \ref{fig:21}, the query encoder is a BiLSTM structure. Assuming the set of all used Chinese characters is $\mathcal{C}$ and its size is $|\mathcal{C}|$, each character $c_i^q$ in query $q$ can be represented as a 1-Hot vector $H(c_i^q)$ with a length of $|\mathcal{C}|$. We initialize the character embedding $E_c$ with a standard normal distribution. The hidden states of character $c_i^q$ are as follows.
\begin{align*}
	\overrightarrow{h_i} & = LSTM(\overrightarrow{h_{i-1}}, H(c_i^q) E_c),
	\\
	\overleftarrow{h_i}  & = LSTM(\overleftarrow{h_{i+1}}, H(c_i^q) E_c),
\end{align*}
$\overrightarrow{h_i}$ is the forward hidden state, while $\overleftarrow{h_i}$ is the backward hidden state. The concatenation $h_i=[\overrightarrow{h_i}; \overleftarrow{h_i}]$ of $\overrightarrow{h_i}$ and $\overleftarrow{h_i}$ is treated as the full hidden state of character $c_i^q$.

\subsection{Feature Encoder}

As for each character $c_i^q$ in query $q$, its feature bag is $F_i$, and $\left \langle L, R \right \rangle \in F_i$ is the features from one context. In feature encoder, because $L$ and $R$ contain the same kinds of features, same network structure can be used to encode them. Fig \ref{fig:22} shows the structure of the feature encoder.


Assuming the window size is $t$, both $L$ and $R$ contain $t$ characters and a distant number $k$. We can use the following structure to accept both left and right features:
\begin{align*}
	e^c = \frac{1} {t} \sum_{c} {H(c)E_c}, \ \ e^d = G(k)E_d.
\end{align*}
$H(c)$ is the 1-Hot vector of character $c$. $E_c$ is the same character embedding that has been used in Query Encoder. We just average the embedding of characters in $\mathcal{C}$ as the representation of character features. $G(k)$ is the 1-Hot vector of distant number $k$ whose dimension depends on the range that $k$ can take. The maximum value of $k$ is not larger than the max length of all segments. $E_d$ is the distance embedding matrix which is initialized by a standard normal distribution.

We concatenate character feature vector $e^c$ and distance feature vector $e^d$ as the whole representation of $L$ or $R$. Then, a linear layer $W$ is applied on $e$ to integrate character and distance features.
\begin{align*}
	e = [e^c; e^d], \ \ g = \tanh(We).
\end{align*}

By applying the above operations to each $\left \langle L_j, R_j \right \rangle \in F_i$, we can get the $g_l$ for left features $L_j$ and $g_r$ for right features $R_j$. The concatenation $f_{j} = [g_l; g_r]$ is treated as the full representation of the $j$th context of character $c_i^q$ in query $q$.
$(f_1, f_2, \ldots, f_j, \ldots)$ are feature vectors of all context in feature bag $F_i$. We use the following formulas to calculate the weights $(\alpha_1, \alpha_2, \ldots, \alpha_j, \ldots)$ where $\alpha_i$ is the weight of $f_i$.
\begin{align*}
	w_j = \tanh(f_j^TU)h_i, \ \ \alpha_j = \frac{\exp{(w_j)}}{\sum_j \exp{(w_j)}},
\end{align*}
where $h_i$ is the hidden state of BiLSTM of the character $c_i^q$ from the query encoder. Because $h_i$ is strongly related to the label of character $c_i^q$, the context feature $f_j$ that is more related to $h_i$ should receive more attention. $U$ is a matrix with size $|f_j| \times |h_i|$.  Finally, the context bag representation $b_i$ is the weighted sum of each $f_j$, $M$ is the context size of the character $c_i^q$.
\begin{align*}
	b_i = \sum_j^M \alpha_j f_j.
\end{align*}

\subsection{Label Decoder}
Label decoder module uses a CRF layer to predict the label sequence $\bm{y} = (y_1,  \ldots, y_i, \ldots, y_n)$, where $y_i$ is the label of $c_i^q$. CRF has been used widely in many sequence labeling tasks. Assuming the input to CRF layer is $\bm{z} = (z_1, \ldots, z_i, \ldots, z_n)$, where $z_i$ can be treated as full vector representation of character $c_i^q$. The conditional probability of any possible label sequence $\bm{y}$ given $\bm{z}$ of query $q$ can be formalized as follow,
\begin{align*}
	& p(\bm{y} | \bm{z}) = \frac{\prod_{i=1}^n \phi(y_{i-1}, y_i, z_i)} {\sum_{y' \in \mathcal{Y}(z)}\prod_{i=1}^n \phi(y_{i-1}', y_i', z_i)}, \\
	& \phi(y', y, z)    = \exp(W_{y', y}^T z).
\end{align*}
Where $\phi(*)$ is exponential function. $W_{y', y}$ is the weight vector corresponding to label pair $(y', y)$. Parameters can be learnt by maximizing the log-likelihood,
\begin{align*}
	L = \sum_{z \ of \ q \in Q} \log p(y|z).
\end{align*}

There are 3 methods to choose vector $z_i$ for character $c_i^q$. The first method is $z_i = h_i$. Only the hidden state of characters in $q$ are used to predict labels, which is the common BiLSTM-CRF model. Because this method only take the query itself into consideration, we call it BiLSTM-CRF(Q). The second method is $z_i = b_i$, which predicts labels mainly based on contexts. This method is called BiLSTM-CRF(C). In the last method, $z_i = [h_i; b_i]$ concatenates the hidden state $h_i$ and context bag information $b_i$. This method called BiLSTM-CRF(Q+C) relies on both the query and contexts.

\section{Experiments}
\label{sec:exp}

\subsection{Datasets}
\begin{figure*}[ht]
\centering
\begin{minipage}[t]{0.48\textwidth}
\centering
	\includegraphics[width=0.65\columnwidth]{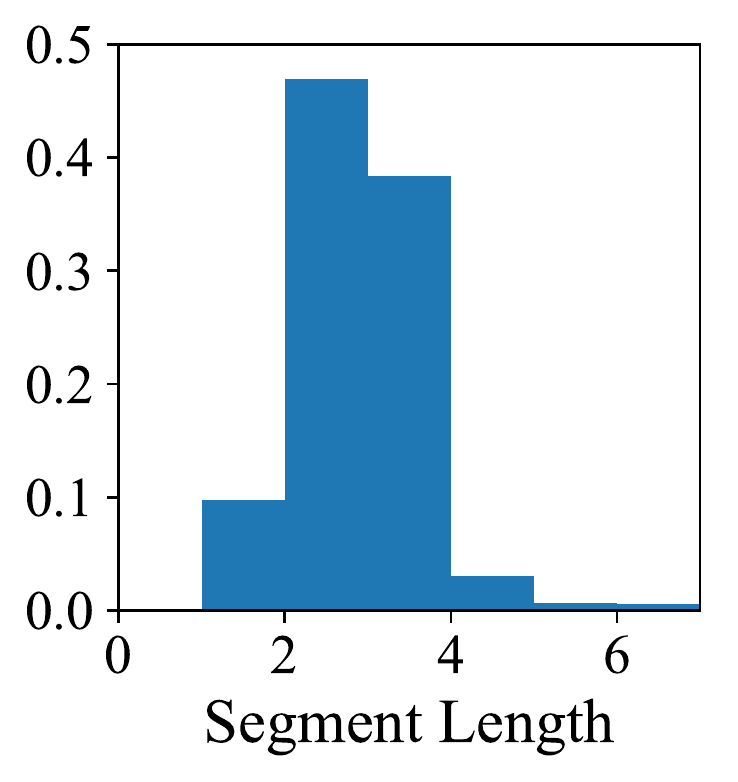}
    \centerline{(a) Distribution of segments.}
	\label{fig:1}
\end{minipage}
\begin{minipage}[t]{0.48\textwidth}
    \centering
	\includegraphics[width=0.65\columnwidth]{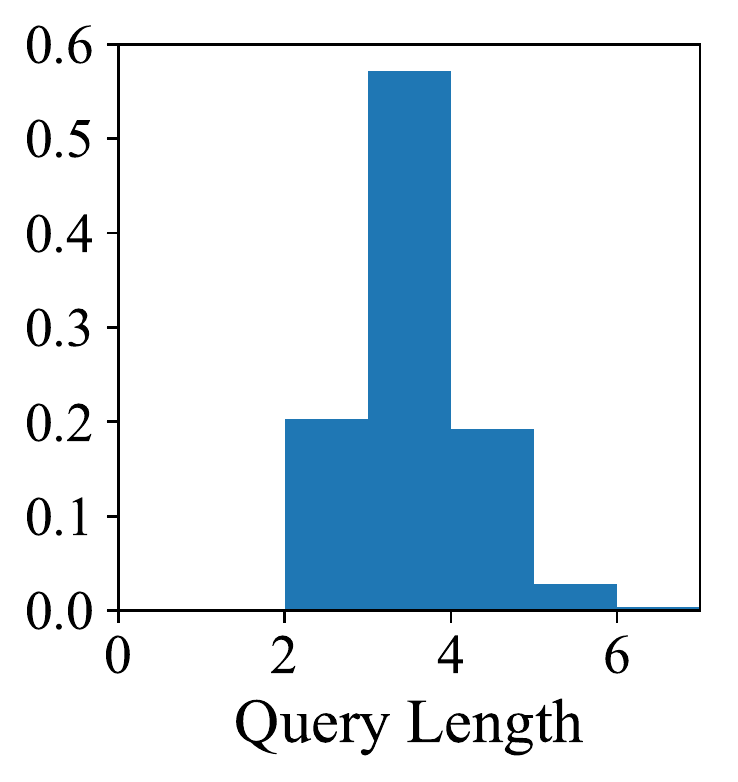}
    \centerline{(b) \small Distribution of queries.}
		\label{fig:2}
\end{minipage}
\caption{Segment and query length distributions of dress dataset.}
	\label{fig:12}
\end{figure*}

We choose two typical categories, dress and bag, and create two datasets. For each dataset, we use the same method to obtain gold segmented queries and external documents, such as brand names, product names, attribute names and attribute values in this category and build a small size dictionary. To ensure the quality of such automatically generated labels, we only reserve queries that all segments of these chosen queries must appear in the dictionary. For the dress dataset, we collect about 25k queries for training and 1.8k queries for test. The number of documents is 10k. For bag category, about 20k and 2k queries are for training and test, respectively. The number of external documents is 15k. Figure \ref{fig:12} shows length distributions of segments and queries using dress dataset as example. Note that the length of a segment is the number of characters in the segment, while the length of a query is the number of segments.

\subsection{Implementation Details}

We select $10\%$ of the queries from the training set as the validation data. All hyper parameters are tuned on the validation data. The character embedding size is tuned from $5$ to $50$ by a step of $5$ each time. The best embedding size is $10$, which seems really small compared with other deep learning models. The reason is that the character set size in our task is also very small. The hidden size of a LSTM cell is set to $10$ too and the size of distance embedding is set to $5$. We use the Adam algorithm with the learning rate $0.0001$ to perform back propagation. The batch size is set to $32$. All parameters are initialized based on a standard normal distribution. The upper limit size of context bag is $5$. In other words, if there are more than $5$ contexts found for a character, we only use $5$ contexts randomly.

We choose two kinds of metrics to evaluate segmentation performance. The first kind of metrics is the commonly used P.R.F. (Precision, Recall and F1) to evaluate the ability to recognize correct segments\cite{shao2017recall}. The second metric is Query Accuracy (QA) \cite{hagen2011query} that is the percentage of queries segmented correctly.

\label{sec:results}

\begin{table*}[ht]
	\centering
 	\caption{Evaluation metrics of all baselines and our models on two datasets.\label{tab:main}}
	\setlength{\tabcolsep}{1.3mm}{
	\begin{tabular}{lcccccccc}
		\toprule
		\multirow{2}{*}{Models} & \multicolumn{4}{c}{Dress} & \multicolumn{4}{c}{Bag}                                                                                                       \\
		\cmidrule(lr){2-5} \cmidrule(lr){6-9}
		                        & P                 & R                  & F1             & QA & P      & R         & F1             & QA \\
		\midrule
		SnowNLP                 & 0.287                     & 0.448                   & 0.350          & 0.074          & 0.242          & 0.403          & 0.303          & 0.116          \\
		THULAC                  & 0.459                     & 0.572                   & 0.509          & 0.291          & 0.385          & 0.488          & 0.431          & 0.333          \\
		Jieba                   & 0.478                     & 0.614                   & 0.538          & 0.275          & 0.469          & 0.586          & 0.521          & 0.387          \\
		\midrule
		UNS                     & 0.427                     & 0.428                   & 0.427          & 0.221          & 0.377          & 0.404          & 0.390          & 0.294          \\
		UNS(-Queries)           & 0.542                     & 0.469                   & 0.503          & 0.326          & 0.350          & 0.385          & 0.366          & 0.268          \\
		UNS(-Documents)         & 0.292                     & 0.234                   & 0.260          & 0.117          & 0.206          & 0.180          & 0.192          & 0.132          \\
		\midrule
		Word2Vec-LR             & 0.687                     & 0.597                   & 0.639          & 0.470          & 0.661          & 0.685          & 0.672          & 0.587          \\
		Perceptron              & 0.650                     & 0.639                   & 0.644          & 0.499          & 0.826          & 0.831          & 0.828          & 0.798          \\
		CRF                     & 0.659                     & 0.642                   & 0.650          & 0.503          & 0.847          & 0.840          & 0.844          & 0.821          \\
		BiLSTM-CRF(Q)           & 0.815                     & 0.792                   & 0.804          & 0.705          & 0.802          & 0.805          & 0.803          & 0.760          \\
		\midrule
		BiLSTM-CRF(C)           & 0.834                     & 0.830                   & 0.832          & 0.732          & 0.706          & 0.774          & 0.739          & 0.661          \\
		BiLSTM-CRF(Q+C)         & \textbf{0.855}            & \textbf{0.851}          & \textbf{0.853} & \textbf{0.759} & \textbf{0.868} & \textbf{0.867} & \textbf{0.867} & \textbf{0.837} \\
		\bottomrule
	\end{tabular}}

\end{table*}

\subsection{Baselines}

There are three kinds of baselines, including the existing tools, unsupervised approaches and supervised approaches. 
We choose three popular and open source CWS tools, SnowNLP\footnote{https://github.com/isnowfy/snownlp}, THULAC \cite{sun2016thulac} and Jieba\footnote{https://github.com/fxsjy/jieba}. As for unsupervised approaches, we implement the model from \cite{risvik_query_2003} called UNS that is based on frequency count and mutual information. UNS is learnt on queries and external documents. UNS(-Queries) and UNS(-Documents) are learnt without queries or external documents, respectively. As for supervised approaches, we choose three existing models, Word2Vec-LR \cite{kale_towards_2017} that is a simple deep learning model based on word embedding, traditional feature-based Perceptron model \cite{du_perceptron-based_2014} and CRF model \cite{yu_query_2009}. BiLSTM-CRF(Q) that only relies on hidden vector of characters in a query is also one of our baselines.

\begin{figure*}[h!t]
\centering
\subfigure[]{\label{fig:curve}\includegraphics[scale=0.35]{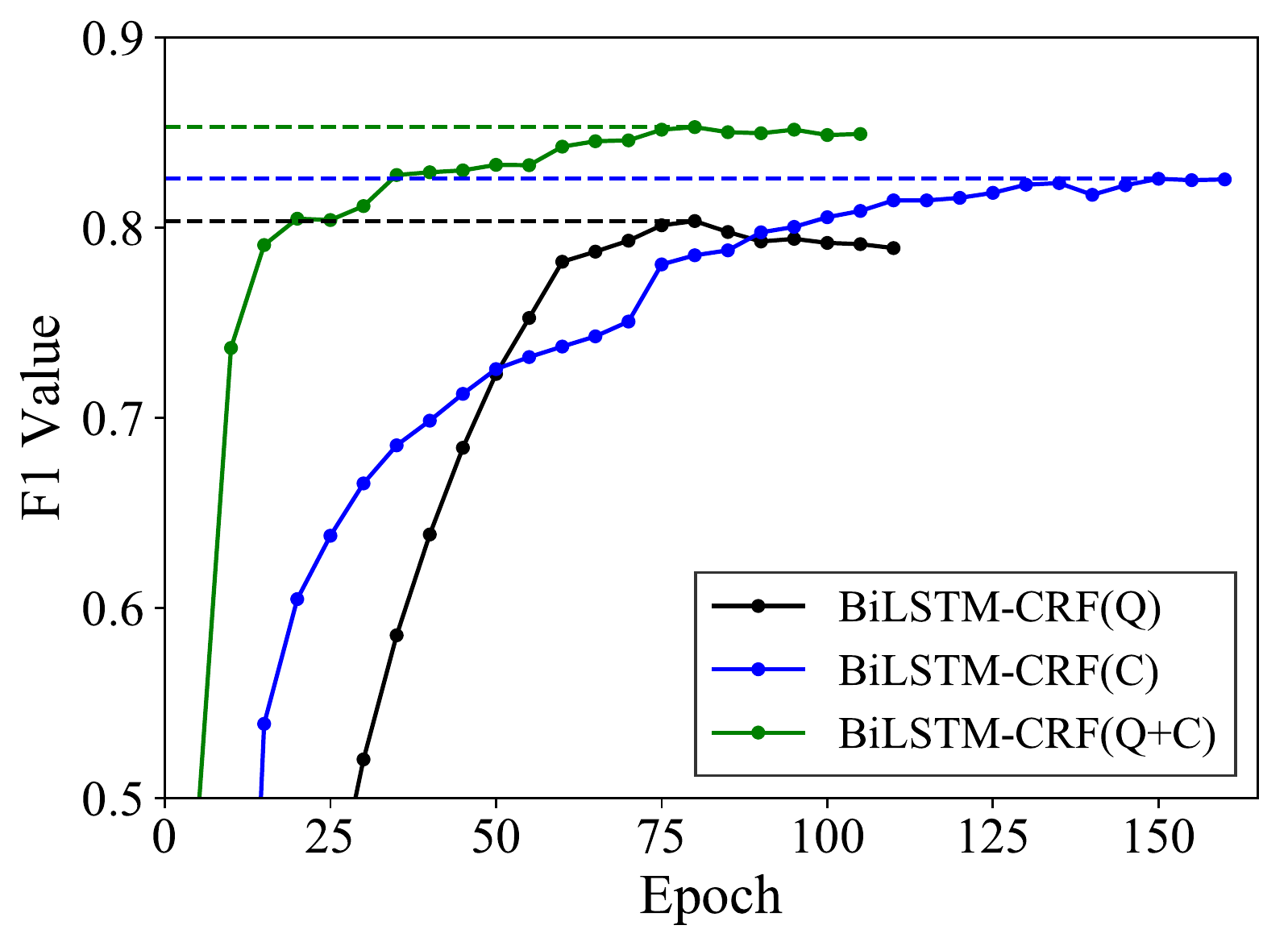}}
\subfigure[]{\label{fig:alpha}\includegraphics[scale=0.35]{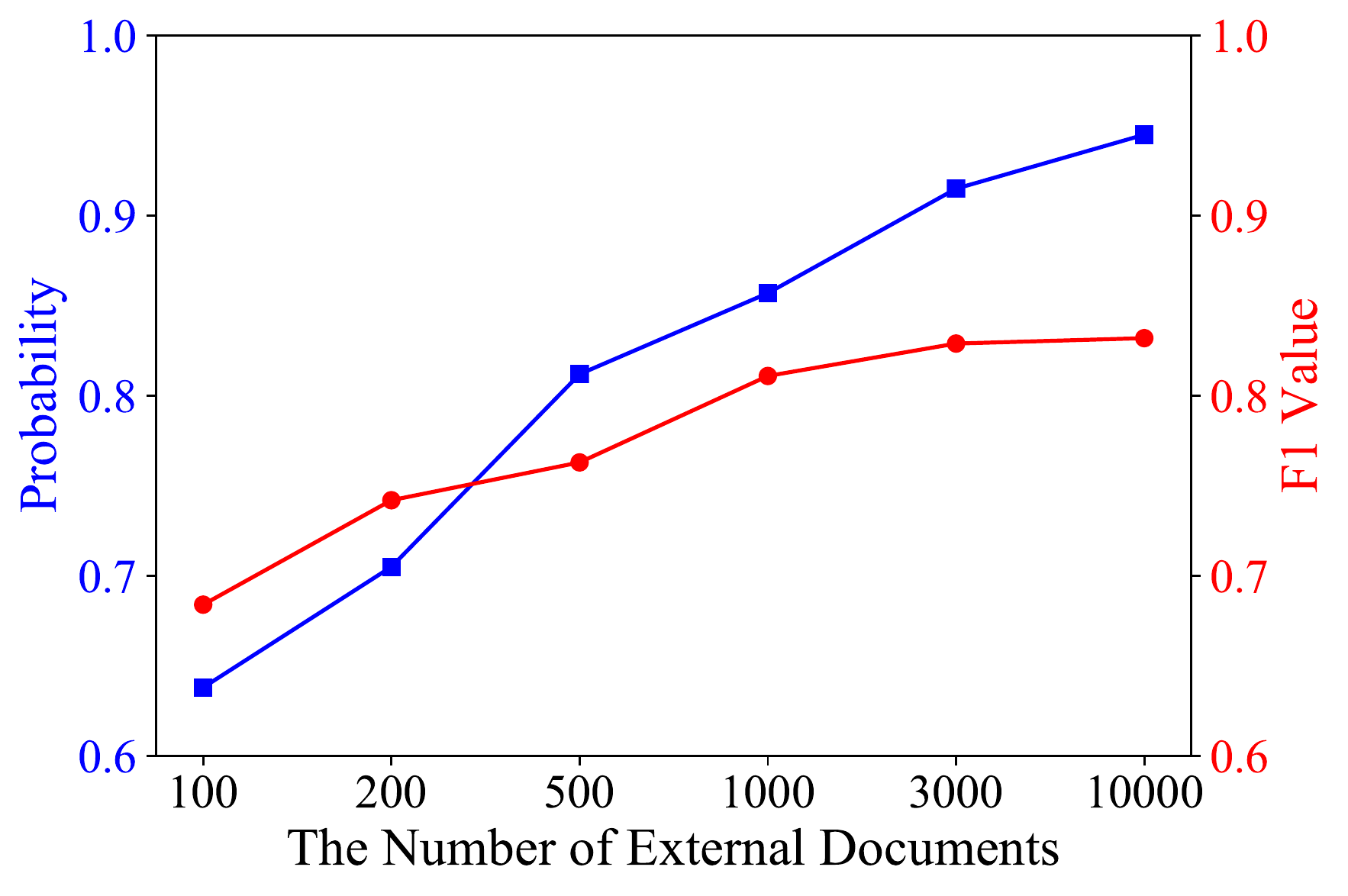}}
\vspace{-1em}
\caption{(a) The learning trends of BiLSTM-CRF(Q), BiLSTM-CRF(C) and BiLSTM-CRF(Q+C). (b) With the increase of the number of external documents, the bule line shows the change of probability that at least 1 contexts can be found, and the red line shows the change of F1 value of BiLSTM-CRF(C).}
\label{fig3}
\end{figure*}

As shown in Table \ref{tab:main}, our BiLSTM-CRF(Q+C) is the best performing model on both the dress and bag datasets. In the dress category, the best baseline is BiLSTM-CRF(Q), and our BiLSTM-CRF(Q+C) achieves 0.049 improvements in F1 value. In the bag category, the best baseline is CRF, and our BiLSTM-CRF(Q+C) beats it by 0.023 in F1 value.

Existing tools trained on open domain sentences for CWS task work rather poorly for query segmentation in e-commerce field. The data distributions of sentences in CWS and queries in query segmentation are very different. Somehow to our surprise, Jieba is better than THULAC in F1 value but worse in Query Accuracy. The reason is that although Jieba can recognize more correct segments, these correct segments are distributed in different queries dispersively. Unsupervised models also do not work well, even worse than the existing Jieba tool. Supervised models trained on labeled queries are much better than other baselines. BiLSTM-CRF(Q) and feature-based CRF are the two best baselines. BiLSTM-CRF(Q) beats CRF by 0.154 on the dress dataset in F1 value, while CRF outperforms BiLSTM-CRF(Q) by 0.041 on the bag dataset.

The effectiveness of our added contexts from external documents can be proved from two aspects. BiLSTM-CRF(C) that predicts labels only on the information from contexts can still work well. BiLSTM-CRF(C) is better than the best baseline on the dress dataset. This means the boundary information in contexts can be used to segment queries independently. From another perspective, BiLSTM-CRF(Q) can work much better after adding contexts. By comparing BiLSTM-CRF(Q+C) with BiLSTM-CRF(Q), the inclusion of contexts brings 0.049 and 0.064 improvements in F1 value on the dress and bag datasets, respectively.

To get more details on the training process, Figure \ref{fig:curve} shows the growth of F1 value of BiLSTM-CRF(Q), BiLSTM-CRF(C) and BiLSTM-CRF(Q+C). Due to early stop strategy, the stop epoch of different models are different.

\section{Related Work}
\label{sec:related2}

Query segmentation task has been studied in research community for a long time. As far as we know, \cite{risvik_query_2003} is the first work which defines such task. They proposed an unsuperised approach based on a score calculated by the frequency count and mutual information (MI). Many following unsupervised approaches \cite{mishra_unsupervised_2011,hagen_power_2010,hagen_query_2011,parikh_segmentation_2013} are similar to \cite{risvik_query_2003} but use different indexes to calculate their scores.  \cite{mishra_unsupervised_2011} calculated scores based on the principal eigenspace similarity of frequency matrix and Hoeffding’s Inequality respectively. \cite{mishra_unsupervised_2011} argues that Hoeffding’s Inequality index can help to detect more rare units than MI approach. \cite{hagen_power_2010,hagen_query_2011,parikh_segmentation_2013} just use n-gram frequencies of the segments of queries in the unsupervised approach. Especially, \cite{tan_unsupervised_2008} train language models using large unlabelled data to do query segmentation.

As for supervised approaches, various kinds of features are designed to train SVM \cite{bergsma_learning_2007}, CRF \cite{yu_query_2009} and Perceptron \cite{du_perceptron-based_2014} to adress query segmentation task. Different from these feature-based approaches, \cite{kale_towards_2017} trains a simple binary classifier to predict segmentation boundaries only based on the word embedding. \cite{lin_query_2017} applies the popular RNN Encoder-Decoder model, which encodes the query into a context vector, and decodes the same query with some special segmentation signs. However, their experiments show this Encoder-Decoder model is not effective at all. Salehi et al.\cite{salehi_multitask_2018} introduce multitask learning for query segmentation. They use the semantic category of the words as an auxiliary task to improve query segmentation task, when the model is also trained to predict the semantic category of the query terms.

\section{Conclusion}
\label{sec:conclusion}


In this paper, we take advantage of external documents to help the query segmentation task. Specifically, for each character, we use its left and right bi-grams to find contexts in external documents. Then, we extract character and distance features from these contexts. We propose an attention network to encode contexts of a character and get a vector representation. This vector is added to the BiLSTM-CRF model to predict character labels. Our BiLSTM-CRF(Q+C) achieves 0.049 and 0.023 improvements in F1 value compared with the existing approaches on both the dress and bag datasets. These results show that contexts in external documents provide highly valuable boundary information to query segmentation task.



%
%
%
\newpage
\bibliographystyle{splncs04}

\bibliography{paper.bib}
\end{CJK}
\end{document}